%% file: acl2023.tex
\pdfoutput=1

\documentclass[11pt]{article}

\usepackage[]{ACL2023}

\usepackage{times}
\usepackage{latexsym}
\usepackage{booktabs}
\usepackage{comment}
\usepackage{amsmath} 
\usepackage{mathtools}
\usepackage{array,multirow,graphicx}
\usepackage{float}
\usepackage{makecell}
\usepackage{amssymb}
\usepackage{pifont}
\usepackage[T1]{fontenc}

\usepackage[utf8]{inputenc}

\usepackage{microtype}

\usepackage{inconsolata}

\usepackage{varwidth}
\newcommand\NameEntry[1]{%
  \multirow{2}{*}{%
    \begin{varwidth}{20 cm}
    \flushleft #1%
    \end{varwidth}}}

\newcommand\NameEntryTwo[1]{%
  \multirow{2}{*}{%
    \begin{varwidth}{10 cm}
    \flushleft #1%
    \end{varwidth}}}

\usepackage{tabularx}
\usepackage{longtable}

\title{What's the Meaning of\\Superhuman Performance in Today's NLU?}

\author{Simone Tedeschi$^{1,2}$, \quad Johan Bos$^3$, \quad Thierry Declerck$^4$, \quad Jan Haji\v{c}$^5$, \\ {\bf Daniel Hershcovich$^6$,} \quad {\bf Eduard H. Hovy$^{7,8}$,} \quad {\bf Alexander Koller$^9$,} \quad {\bf Simon Krek$^{10,11}$,} \\ \quad {\bf Steven Schockaert$^{12}$,} \quad {\bf Rico Sennrich$^{13,14}$,} \quad {\bf Ekaterina Shutova$^{15}$,} \quad {\bf Roberto Navigli$^2$} \\
$^1$Babelscape \quad $^2$Sapienza University of Rome \quad
$^3$University of Groningen \\ $^4$German Research Center for AI (DFKI) \quad 
$^5$Charles University \quad
$^6$University of Copenhagen \\ $^7$University of Melbourne \quad $^8$Carnegie Mellon University \quad
$^9$Saarland University \\
$^{10}$Jožef Stefan Institute \quad $^{11}$University of Ljubljana \quad 
$^{12}$Cardiff University \\ $^{13}$University of Zurich \quad
$^{14}$University of Edinburgh \quad
$^{15}$University of Amsterdam\\
\texttt{\{tedeschi, navigli\}@diag.uniroma1.it}, \quad \texttt{johan.bos@rug.nl} \\
\texttt{declerck@dfki.de} \quad \texttt{hajic@ufal.mff.cuni.cz} \quad \texttt{dh@di.ku.dk}\\
\quad \texttt{hovy@cmu.edu} \quad \texttt{koller@coli.uni-saarland.de} \quad \texttt{simon.krek@ijs.si} \\
\texttt{schockaerts1@cardiff.ac.uk} \quad \texttt{sennrich@cl.uzh.ch} \quad \texttt{e.shutova@uva.nl}
}

\begin{document}

\maketitle

\begin{abstract}
In the last five years, there has been a significant focus in Natural Language Processing (NLP) on developing larger Pretrained Language Models (PLMs) and introducing benchmarks such as SuperGLUE and SQuAD to measure their abilities in language understanding, reasoning, and reading comprehension. These PLMs have achieved impressive results on these benchmarks, even surpassing human performance in some cases. This has led to claims of superhuman capabilities and the provocative idea that certain tasks have been solved. 
In this position paper, 
we take a critical look at these claims and ask whether PLMs truly have superhuman abilities and what the current benchmarks are really evaluating. We show that these benchmarks have serious limitations affecting the comparison between humans and PLMs and provide recommendations 
for fairer and more transparent benchmarks. 
\end{abstract}

\input introduction

\input overview

\input leaderboards
\input role-of-data

\input role-of-annotators


\input explanations

\input recommendations

\input conclusions

\input limitations

\input{acknowledgements}


\bibliography{anthology,custom}
\bibliographystyle{acl_natbib}

\appendix

\input appendix-errors



\end{document}

%% file: introduction.tex
\section{Introduction}

\begin{figure}
\resizebox{\columnwidth}{!}{%
\centering
\includegraphics{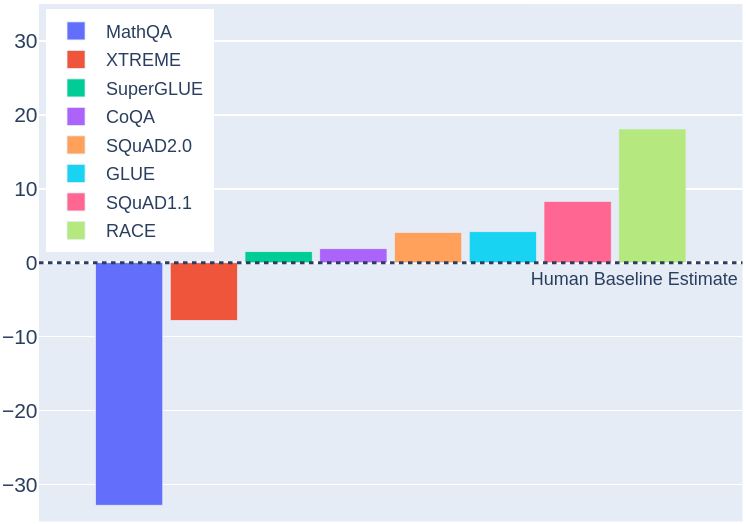}
}
\caption{Difference between the scores obtained by the best-performing systems and humans in various popular NLP benchmarks. 
The systems outperform humans on 6 out of 8 of the reported benchmarks (best seen in color).
}
\label{fig:bar-plot}
\end{figure}

In recent years, research in the field of Natural Language Processing (NLP) has been driven by a frantic race to reach the top spot in popular benchmarks \cite{wang-etal-2018-glue, wang-et-al-2019-superglue, lai-etal-2017-race, rajpurkar-etal-2018-know, reddy-etal-2019-coqa}.
Typically the race takes the shape of a rapid cycle of parameter tuning updates by several teams, communicating their results using a shared leaderboard. 
Not infrequently, systems achieve better-than-human performance on several tasks (see Figure~\ref{fig:bar-plot}).  
Yet what does this level of performance really mean for NLP?  
The impressive capabilities of ChatGPT make this question even more urgent. 

It is relatively easy to outperform humans with simple procedural tasks like arithmetic and extreme memory-intensive tasks involving vast amounts of data. 
But most tasks involving natural language typically require knowledge and inference.  Do high-performing NLP algorithms really have (super)human capabilities? Or are the metrics that deliver these scores suspect?  

Given the impact of claiming superhuman performance, it is important for researchers to understand exactly what is going on. 
As many in NLP have experienced, the false sense of accomplishment of superhuman performance often leads to an abrupt disappointment when a supposedly superb system is applied to realistic data in a real-world situation.  By propounding unrealistic claims, NLP researchers harm themselves and the field as a whole.  

Some problems result from the metrics used to assess systems, which are invariably automated, and the data these metrics employ, which may be skewed in various ways. The metrics might give incomplete or biased reports of performance, or simply not apply in certain situations. 

Other problems arise from the `boundary parameters' that shape the task, which are usually not adequately reflected in the evaluation metric, very seldom in the kinds of automated metrics used in leaderboards. Specifically, the correctness of a task setup and its dataset instances should not be taken for granted. Also, humans and machines are often evaluated under different conditions, such as the level and type of knowledge provided to perform the task and the test items used to compute performance. 

Yet other problems result from the nature of leaderboard-based evaluation. 
Despite the obvious benefit of driving development through competition with little human effort, these evaluations typically do not foster understanding. Teams driven by a rapid evaluation turnaround cycle in a competitive mode tend to focus more on quantitative results than on error analyses which aim at improving awareness of their problem. 
As currently set up, benchmarks and comparisons do not incentivize a deeper understanding of the systems' performance, nor do they foster research geared towards producing automatic explanations: it is one thing to produce a numerical system performance score, but quite another to rate the adequacy and understandability of an explanation.  

In this paper, we explore the interpretation of the
superhuman performance and the utility of leaderboards, discuss how human performance is actually computed in a range of tasks, and how requirements differ for humans and automatic systems across tasks. We hope to encourage leaderboard creators to be more circumspect when setting up their challenges and provide clear `boundary conditions' and descriptions of the limitations of their evaluations.

%% file: overview.tex
\section{Popular Leaderboards Are Saturated}\label{sec:overview}

Leaderboard-based evaluation has become a popular practice in NLP\footnote{We mainly focus on leaderboards for Natural Language Understanding (NLU) due to their commonalities, but note that Natural Language Generation (NLG) tasks such as machine translation, where the most accepted `leaderboards' are based on human scoring, have their own set of issues \citep[e.g.][]{laeubli2020}.}
\cite{wang-etal-2018-glue, wang-et-al-2019-superglue, lai-etal-2017-race, rajpurkar-etal-2018-know, reddy-etal-2019-coqa}. The goal of these leaderboards is to encourage the development of systems capable of solving certain tasks and to measure their progress by comparing the best systems against humans. Their great success has led many researchers to focus on just the proposed tasks, resulting in a rapid saturation of the scores which, in many tasks, are equal to or greater than those obtained by humans. As a consequence, many have attributed superhuman performance to such systems, and some tasks have been deemed solved. However, while systems in some areas of AI are compared with the best possible humans, e.g.\ IBM Deep Blue vs.\ Garry Kasparov in chess\footnote{\url{https://en.wikipedia.org/wiki/Deep_Blue_versus_Garry_Kasparov}} or IBM Watson vs.\ Ken Jennings and Brad Rutter in the \textit{Jeopardy!} quiz show\footnote{\url{https://en.wikipedia.org/wiki/IBM_Watson\#Jeopardy!}}, NLP researchers often naively or vaguely estimate the ``human baseline'', assuming it is a uniform and accepted term of comparison, an established level that systems need to simply beat.  
In this section we provide a broad overview of existing
NLP benchmarks, with a particular focus on NLU leaderboards where human baselines are outperformed by systems, and then show that 
the construction of such benchmarks is fraught with inconsistencies. 

The SuperGLUE benchmark \cite{wang-et-al-2019-superglue} is a well-known framework for evaluating research towards general-purpose language understanding models for English. It is a collection of 10 language understanding tasks built on existing public datasets, together with private test data, an evaluation server, and a single-number performance metric. In many tasks, humans are outperformed by the best-scoring systems, often by a large margin, ranking 8th in the current overall leaderboard. Likewise, the SuperGLUE predecessor, i.e.\ GLUE \cite{wang-etal-2018-glue}, was built to measure advances in NLU, and the systems' scores quickly saturated the benchmark, thereby sliding the human baseline down to the 23rd position in the ranking.

The RACE benchmark \cite{lai-etal-2017-race} was designed specifically for evaluating NLP models on a set of challenging reading comprehension tasks, such as Question Answering (QA) and text summarization. It consists of a large dataset of more than 28,000 multiple-choice questions, which are drawn from middle and high school problems extracted from English
examinations in China. These questions cover a wide range of topics and require the ability to reason, understand context, and make inferences based on the provided text. Human baselines rank 21st on the public leaderboard, with a gap of almost 20 points compared to the best-scoring system. Similarly, the SQuAD2.0 benchmark \cite{rajpurkar-etal-2018-know} is another popular collection of reading comprehension questions and answers based on Wikipedia articles. The questions are created by crowdworkers and the answers are a portion of text from the corresponding article. The peculiar difference of this benchmark compared to SQuAD1.1 \cite{rajpurkar-etal-2016-squad} is that some of the questions may not have answers, hence systems are required to learn to abstain as well. Again, the human baseline is placed in low positions of the ranking, reaching just the 30th place. Another notable, related benchmark is CoQA \cite{reddy-etal-2019-coqa}, a large-scale dataset focused on Conversational QA systems. 
In this task, humans rank 7th, with a gap of 2 points from the top system.

Quite different results are observed when moving to a cross-lingual scenario or when systems are required to perform mathematical and logical reasoning. In particular, XTREME \cite{hu-etal-2020-xtreme} is a benchmark for cross-lingual transfer evaluation that covers dozens of languages spanning 12 language families, and includes 9 tasks that require reasoning about different levels of syntax or semantics. In this case, the human baselines beat the systems in all tasks, with an overall score 8 points higher than that of the best-performing system. XTREME has been succeeded by XTREME-R \cite{ruder-etal-2021-xtreme}, which covers 14 language families and includes 10 tasks. Similarly, when systems are evaluated over MathQA \cite{amini-etal-2019-mathqa} inputs, i.e.\ mathematical questions in the form of natural language, systems perform poorly compared to humans. Indeed, humans achieve an accuracy of 87\% while systems only reach 54.2\%. Since systems are still far from human-level performance in these benchmarks, they are out of the scope of our study. However, the highlighted gaps should encourage further research in these areas.

An alternative view on system evaluation is presented by the adversarial evaluation framework \cite{nie-etal-2020-adversarial,kiela-etal-2021-dynabench}, where the evaluation is performed through an iterative ``human-and-model-in-the-loop'' annotation process. Humans are asked to inspect the model output and produce adversarial examples that target specific model weaknesses. The evaluation target is thus a moving goalpost, as opposed to the static targets of most other benchmarks, which saturate quickly. The Dynabench benchmark \cite{kiela-etal-2021-dynabench} embraces this approach, incorporating tasks such as NLI, QA, sentiment analysis and hate speech detection. It provides a platform for the annotators to examine model output and create adversarial examples. At the time of writing, most of the tasks within Dynabench do not report human performance, however. Exceptions include the adversarial visual QA task \cite{AdversarialVQA}, where the proposed adversarial examples are solved by other humans and agreement is computed in terms of accuracy. Model performance in this setting falls far below the human performance.

Using more challenging examples for model evaluation, and possibly subsequent re-training, is an appealing approach, likely to strengthen the models with respect to the aspects that the examples target. The caveat is, however, that special care needs to be taken to avoid loss of generality. The annotation of adversarial examples directly  depends on the behavior of the model (or set of models) 
under consideration; the addition of a large number of adversarial examples will likely change the data distribution by potentially overemphasizing rare events; finally, the annotators may focus on a small number of properties of the model, thus ``overfitting'' the models. 

Although there are many other popular NLP benchmarks to be investigated, e.g. XGLUE \cite{liang-etal-2020-xglue} and SentiBench \cite{ribeiro2016sentibench}, we limit our review to those in which human performance is provided and that can therefore help us answer the main question of this paper concerning the meaning of superhuman performance.


%% file: leaderboards.tex
\section{Human Baselines Are Not Reliable}\label{sec:lea}

As discussed above, 
 many NLU benchmarks are saturated (cf.\ Figure~\ref{fig:bar-plot}). Here we dive deeper into some of them, 
 identify the reasons for their quick saturation, 
 and discuss whether it is fair to claim superhuman performance of state-of-the-art models.
In particular, we study SuperGLUE \cite{wang-et-al-2019-superglue} and SQuAD \cite{rajpurkar-etal-2016-squad, rajpurkar-etal-2018-know}, as the representatives for general language understanding and reading comprehension, respectively.

\subsection{SuperGLUE}

For each of the ten tasks in SuperGLUE, human performance is provided and systems are compared against it. Specifically, for four of these tasks -- Word in Context \cite[WiC,][]{pilehvar-camacho-collados-2019-wic}, Multi-Sentence Reading Comprehension \cite[MultiRC,][]{khashabi-etal-2018-looking}, Recognizing Textual Entailment
\cite[RTE,][]{nangia-bowman-2019-human}, Reading Comprehension with Commonsense Knowledge \cite[ReCoRD,][]{zhang-et-al-2018-record} -- human performance is computed by the authors of the corresponding papers, while for the remaining tasks\footnote{Boolean Questions (BoolQ), Commitment Bank (CB), Choice of Plausible Alternatives (COPA), Winograd Schema Challenge (WSC), Broadcoverage Diagnostics (AX-b), Winogender Schema Diagnostics (AX-g).} humans are evaluated by the creators of the SuperGLUE benchmark. 

\begin{table*}[ht!]
\addtolength{\tabcolsep}{10pt}
\centering
\resizebox{\linewidth}{!}{%
\begin{tabular}{@{}lccccccccccc@{}}
\toprule
\textbf{Model} & \textbf{BoolQ} & \textbf{CB} & \textbf{COPA} & \textbf{MultiRC} & \textbf{ReCoRD} & \textbf{RTE} & \textbf{WiC} & \textbf{WSC} & \textbf{AX-g} & \textbf{AX-b}\\
\toprule
\textsc{Vega v2} & 90.5 & \textbf{98.6} & 99.4 & 88.2 & 94.4 & \textbf{96.0} & 77.4 & 98.6 & \textbf{100.0} & \,-0.4\\
\textsc{ST-MoE-32B} & \textbf{92.4} & 96.9 & 99.2 & \textbf{89.6} & 95.1 & 93.5 & 77.7 & 96.6 & 96.1 & 72.3\\
\textsc{Turing NLR v5} & 92.0 & 95.9 & 98.2 & 88.4 & \textbf{96.4} & 94.1 & 77.1 & 97.3 & 93.3 & 67.8\\
\textsc{ERNIE 3.0} & 91.0 & \textbf{98.6} & 97.4 & 88.6 & 94.7 & 92.6 & 77.4 & 97.3 & 92.7 & 68.6\\
\textsc{PaLM 540B} & 91.9 & 94.4 & 99.0 & 88.7 & 94.2 & 94.1 & 77.4 & 95.9 & 95.5 & 72.9\\
\midrule
\textsc{Human Baselines} & 89.0 & 95.8 & \textbf{100.0\ \ } & 81.8 & 91.7 & 93.6 & \textbf{80.0} & \textbf{100.0\ \ } & 99.3 & \textbf{76.6}\\
\bottomrule
\end{tabular}%
}
\caption{\label{tab:superglue} Results on SuperGLUE (from \url{https://super.gluebenchmark.com/leaderboard}). On top, we report the results of the 5 best-performing models, on average, while on bottom we report those of the human baselines on the various tasks. We mark in bold the best scores per task. AX-g and AX-b are the two diagnostic datasets.}
\end{table*}

\paragraph{WiC}\label{wic} For this lexical-semantic task, four sets of 100 instances with an overlap of 50 instances between two of the annotators were randomly sampled from the test set. Each set was then assigned to an annotator, resulting in a total of 300 annotated instances. The annotators were not lexicographers and were not provided with sense distinctions to resemble the more difficult scenario for unsupervised models (cf. Appendix \ref{app:copenhagen-experiment}). A final score of 80\% was then obtained by averaging the individual scores achieved by the humans on the 4 sets (between 79\% and 82\%).

\paragraph{MultiRC}

In the Multi-Sentence Reading Comprehension task, four native-speaker annotators tagged the entire test set of 166 instances. Human performance was obtained by combining the individual predictions of the different annotators via majority voting. 

\paragraph{RTE}
To establish the human performance on the RTE task, annotators were hired through the Hybrid 
data collection platform. Each annotator first completed a short training procedure, during which they were provided with task-specific guidelines and annotated 20 random examples from the dev set.
Only annotators with $\ge 65\%$ accuracy qualified for the main task.
500 examples were randomly taken from the test set and, for each instance, the final label was obtained by combining 5 different annotations via majority voting, reporting a final accuracy of 93.6\%.
The average pay rate 
was \$17/hr for the main task, and  \$7.6/hr for training.

\paragraph{ReCoRD}
For the Reading Comprehension with Commonsense Knowledge task, 2,257 crowdworkers were hired through the Amazon Mechanical Turk platform (AMT).
For first-time workers, the HIT\footnote{A Human Intelligence Task, or HIT, is a question that needs an answer. A HIT represents a single, self-contained, virtual task that a worker can work on, submit an answer, and collect a reward for completing.}
assignments
were accompanied with guidelines.
Crowdworkers were required to have $\ge 50$ HITs with a 95\% HIT acceptance rate and to be located in the USA, Canada, or UK. 
The average pay rate was \$3.6/hr. 

\paragraph{Other SuperGLUE Tasks}
For the six remaining tasks, the SuperGLUE authors hired crowdworkers through AMT: the annotators first completed a short training phase where 30 random development set examples were provided for each task. 
Only workers who completed 5 HITs during training with performance at, or above, the median across all workers were admitted to the main task. Human performance was estimated on a random set of 100 test samples from each task, by applying majority voting on the annotations of 5 workers.
During both phases, workers had access to task-specific guidelines, with a pay rate of  \$23.75/hr.

\subsection{SQuAD}
In SQuAD1.1 \cite{rajpurkar-etal-2016-squad}, the researchers obtained 
$\ge 3$ answers from human workers for each question in the dev and test sets, and estimated human performance by using only one of the answers as the ``human prediction'' and the remaining answers as ``ground truth'' for comparison. Specifically, workers were shown the questions and relevant paragraphs of an article and were asked to select the shortest paragraph span 
that answered the question. 
They were advised to complete 5 questions in 2 minutes with a \$9/hr pay rate.

In SQuAD2.0 \cite{rajpurkar-etal-2018-know}, instead, the authors collected
multiple answers for each question (i.e.\ 4.8 answers, on average) and selected the final human prediction by majority voting. The answers were collected by providing annotators with a paragraph and its associated questions -- unanswerable and answerable ones shuffled together -- and asking them either to highlight the answer in the paragraph or to mark the question as unanswerable. They were asked to spend one minute per question with a \$10.50/hr pay rate.

\subsection{Issues}
\label{sec:issues}
Comparing the performance of the five best systems against humans on SuperGLUE (Table~\ref{tab:superglue}), it is immediately apparent that the machines outperform humans on 6 out of 10 tasks, and often by a large margin (e.g.\ 7.8 F$_1$ points on MultiRC). 
Similarly, best systems substantially outperform humans on SQuAD1.1 and SQuAD2.0, with a margin of 8.3 and 4.1 points in exact match accuracy, respectively.
Interestingly, (zero-shot) ChatGPT performs poorly compared to both human baselines and best-performing (fine-tuned) systems. Indeed, compared to the scores reported in Table~\ref{tab:superglue}, it achieves just 86.8 on BoolQ, 89.3 on CB, 58.0 on COPA, 85.2 on RTE and 64.6 on WiC as measured by \citet{qin-etal-2023-is-chatgpt} and \citet{kocon-etal-2023-chatgpt-jack}. Additionally, \citet{kocon-etal-2023-chatgpt-jack} also showed that ChatGPT performs 20\% worse than state-of-the-art systems on the SQuAD2.0 benchmark, and demonstrated that it is, on average, 25\% worse than specialized ML systems on a wide array of tasks. Hence it is not relevant for our study as its performance is still far from human-level. 

What does appear relevant, instead, are the extremely high, often superhuman, scores achieved by specialized systems. Nevertheless, notwithstanding such scores, in the above-mentioned benchmarks multiple factors make human-to-system comparisons unfair because they limit human performance while facilitating systems. We list them in the remainder of this section.

\vspace{-0.1cm}

\paragraph{Apples and oranges} 
The most glaring problem is that, on almost all SuperGLUE tasks, humans and systems are evaluated on different test sets (i.e.\ on a small subset vs. the full test set). Specifically, in the WiC and RTE tasks, humans are assessed on 21.4\% and 16.6\% of the test set (i.e.\ 300 out of 1400 and 500 out of 3000 instances), respectively. 
Similarly, in the other SuperGLUE tasks humans are evaluated on a subset of 100 instances per task, which -- in the worst case of the BoolQ dataset -- amounts to just 3\% of the test set. We provide more details in Appendix \ref{app:apples-and-oranges}. 

\vspace{-0.1cm}
\paragraph{Human evaluation metrics} 
Different metrics are used to assess humans across tasks. While most of the tasks employ majority voting, WiC merely averages the scores achieved by humans on 4 small distinct subsets. In SQuAD1.1, humans are evaluated by comparing the tags of an arbitrary annotator against those of two other ``ground truth'' annotators, thereby likely underestimating the final score. 
 


\vspace{-0.1cm}
\paragraph{Heterogeneous and unknown pay rates}
Pay rates varied considerably across the various tasks, ranging from undeclared pay rates to \$23.75/hr.
Low and mediocre wages, 
as in ReCoRD and SQuAD, 
may have contributed to suboptimal human performance:  
the \$3.6/hr pay rate on  ReCoRD could be one of the reasons for the large gap between systems and humans, while the unknown pay rate for MultiRC might explain 
the 18.2\% human error rate on this binary classification task.

\vspace{-0.1cm}
\paragraph{Ground-truth data quality} We identified several errors and ambiguous instances in the gold-standard datasets, some of which we report in Table~\ref{tab:errors}. 
Importantly, we note that, while systems can find spurious correlations between training and evaluation instances, and therefore provide the correct answer without clear evidence, 
humans cannot find such correlations, or otherwise may genuinely disagree on what the correct answer is. 
We elaborate on this point in Appendix \ref{appendix:ground-truth-quality}, by analyzing several examples per task, as well as in Appendix \ref{app:copenhagen-experiment}, where we report the results of an ad hoc study concerning the WiC dataset.

\begin{table*}[htbp]
\resizebox{\linewidth}{!}{%
\begin{tabular}{ll}
\toprule
\parbox[t]{2mm}{\multirow{6}{*}{\rotatebox[origin=c]{90}{\textbf{BoolQ}}}} & 
\NameEntry{\textbf{Passage:} \textit{Shower gel -- Shower gels for men may contain the ingredient menthol, which gives a cooling and stimulating sensation on the skin, and some men's shower gels are also designed specifically for use on hair and body. Shower gels contain milder surfactant bases than shampoos, and some also contain gentle conditioning agents in the formula. This means that shower gels can also double as an effective and perfectly acceptable substitute to shampoo, even if they are not labelled as a hair and body wash. Washing hair with shower gel should give approximately the same result as using a moisturising shampoo.}}\\
\\
\\
\\
\\
& \textbf{Question:} \textit{is it bad to wash your hair with shower gel} \quad \textbf{Answer:} \textsc{True}\\
\toprule
\parbox[t]{2mm}{\multirow{3}{*}{\rotatebox[origin=c]{90}{\textbf{CB}}}} & 
\NameEntry{\textbf{Premise:} \textit{A: I do too, so she couldn't possibly turn them out like some of these popular writers, B: Huh-uh. A: but oh, her books are just incredible. I don't think they've ever made a movie, do you?}}\\
\\
& \textbf{Hypothesis:} \textit{they've ever made a movie} \quad \textbf{Entailment:} \textsc{False}\\
\toprule
\parbox[t]{2mm}{\multirow{5}{*}{\rotatebox[origin=c]{90}{\textbf{MultiRC}}}} & 
\NameEntry{\textbf{Paragraph:} \textit{What causes a change in motion? The application of a force. Any time an object changes motion, a force has been applied. 
[\dots] It depends on the strength of the force. It also depends on the objects mass. Think about some simple tasks you may regularly do. You may pick up a baseball. This requires only a very small force.}}\\
\\
\\
& \textbf{Question:} \textit{What factors cause changes in motion of a moving object?} \quad \textbf{Candidate Answers}: \textit{Shape of the object} (\textsc{False}), \\ & \textit{Mass of the object} (\textsc{True}), \textit{The object's mass} (\textsc{False}), $\dots$\\
\toprule
\parbox[t]{2mm}{\multirow{2}{*}{\rotatebox[origin=c]{90}{\textbf{RTE}}}} & 
\textbf{Premise:} \textit{In most Pacific countries there are very few women in parliament.}\\
& \textbf{Hypothesis:} \textit{Women are poorly represented in parliament.} \quad \textbf{Entailment:} \textsc{True}\\
\toprule
\parbox[t]{2mm}{\multirow{2}{*}{\rotatebox[origin=c]{90}{\textbf{WiC}}}} & 
\textbf{Context 1:} \textit{The senator received severe \underline{criticism} from his opponent.}\\ & \textbf{Context 2:} \textit{The politician received a lot of public \underline{criticism} for his controversial stance on the issue.} \quad \textbf{Sense Match:} \textsc{False}\\
\bottomrule
\end{tabular}%
}
\caption{Some errors and ambiguous instances we have found in the gold standard datasets of various SuperGLUE tasks. 
We limited our analysis to tasks where suspiciously low human performance was reported (cf. Table \ref{tab:superglue}).}\label{tab:errors}
\end{table*}

\vspace{-0.1cm}

\paragraph{Information about annotators and instructions} 
Details of the annotator pool (e.g. the number of annotators, their background and nationality, etc.) are often omitted. 
Similarly, the absence of training instructions and task guidelines raises questions about the quality of the training phase, if any. 

%% file: role-of-data.tex
\section{Setups Favor Misleading Comparisons}
\label{sec:role-of-data}


Summarizing the above observations, we find four main sources of human-to-system comparison error. These correspond to the following key aspects of the evaluation process: system performance, the evaluation data, the measurement process, and humans themselves.  
We discuss each in turn.  




\subsection{Systems: Right for the Wrong Reasons}\label{sec:right-for-wrong}

Across a variety of tasks, \citet{sogaard-etal-2021-need} report that random train-test splits consistently overestimate model performance: randomization at the sentence level reduces discrepancies between training and test sets as sentences from the same documents occur in both. Non-random standard splits also bring the danger of inadvertent, community-wide overfitting \citep{gorman-bedrick-2019-need}\footnote{Even with hidden test sets, such overfitting could happen via publication bias or just the faster spread of methods used in ``state-of-the-art'' systems (as measured on a static test set).}. 

In natural language inference (NLI), multiple authors have found that BERT achieves what looks like near-human accuracy by exploiting idiosyncrasies of the data: they are ``right for the wrong reasons'' \cite{mccoy-etal-2019-right,niven-kao-2019-probing}. Here much of BERT's success is attributed to its ability to learn syntactic and lexical cues for inference, which happen to be mostly correct on the original test data. However, these cues do not actually support such inferences on adversarial datasets, taking BERT's accuracy to chance level or below.

\citet{poliak-etal-2018-hypothesis} report an even more extreme case of being ``right for the wrong reason'': 
several NLI datasets support what they call hypothesis-only models, which perform surprisingly well without exposure to the premise \cite{DBLP:journals/corr/abs-1803-02324}, e.g. outperforming the majority-class baseline. 
\citet{poliak-etal-2018-hypothesis} attribute this to statistical irregularities in the data (often single words indicating negation), caused by obvious annotation strategies chosen by 
crowdworkers who were not stimulated enough to come up with more creative ways to produce contradictions or entailments. 
Along the same lines, \citet{parmar-etal-2023-dont}
recently identified instruction bias in 14 NLU benchmarks. Specifically, they found that this phenomenon is evident in most of these datasets, showing
that $\sim$73\% of instruction examples, on average,
share a few clear bias patterns, and that models often fail to generalize beyond such patterns.

\subsection{Data: Monolithicity Obscures Details}\label{sec:monolothic}

A further cause of 
systematic performance overestimation
is that test sets include instances with varied, often unfathomable, levels of difficulty, so 
the exact reported accuracy will be a weighted average that depends 
directly on the mixture of easy and hard instances in the test data. The composition of train-test splits can thus make a big difference \cite{swayamdipta-etal-2020-dataset}.

In QA, \newcite{lewis-etal-2021-question} investigated the train-test splits of several popular datasets. They found that there can be substantial overlap between the answers and even the questions of the training and test sets. The evaluation results differed greatly between seen and unseen questions and answers; for instance, the exact-match accuracy of BART as a closed-book QA system on Web\-Questions dropped from 76\% to below 2\% when neither the question nor the answer were ever seen during training.

In semantic parsing, seq2seq models such as BART and T5 are very accurate when evaluated in-domain on broad-coverage parsing tasks, e.g.\ \newcite{bevilacqua-etal-2021-one}. \newcite{yao-2022-structural-generalization} report that their accuracy drops to close to zero on test subsets that require them to generalize to language that is structurally more complex than the training data. This is corroborated when constructing hard sets, i.e.\ train-test splits based on compositional generalization, forcing the accuracy of seq2seq models below 20\% \cite{bogin-2022-local-structure}.

\subsection{Measurement: Automation Is  Limiting}
\label{sec:automatic-evaluation}

A third key limitation of current evaluations, and especially existing leaderboards, is that they assume that the performance of a model can be measured automatically. 
While this has not been discussed very much in NLU,
in other communities it has long been recognized that automatic evaluations are imperfect proxies of human judgments \cite{novikova-etal-2017-need}. Machine translation papers report BLEU scores because they are drastically cheaper to calculate than the cost to collect human judgments about the fluency and adequacy of text; but one system that outperforms another on BLEU is not necessarily judged better by humans \cite{callison-burch-etal-2006-evaluating,popel_transforming_2020}. 
While recent automatic metrics 
correlate better with human judgments \citep{kocmi-etal-2021-ship}, automatic evaluation has consistently been found 
problematic when comparing top-performing systems \citep{ma-etal-2019-results}. 
%
Similarly, \newcite{byron-etal-2009-report} recommend crowdsourced evaluations 
to counter the inadequacy of automated evaluation for NLG. 

The deeper issue with our reliance on automated evaluations is that they constrain the tasks on which we can evaluate systems. 
New shared tasks and datasets are specifically designed to make automated evaluations possible.
However, many skills that show competent language use cannot easily be 
approximated by automatic measures \cite{dunietz-etal-2020-test}: 
there are entire facets of language competence that are systematically out of scope for the tasks we design. One might argue that these are the most interesting parts of the actual mastery of language.
Therefore, 
human-level performance on automatically-evaluated tasks does not equate to human-level performance on real language use.

%% file: role-of-annotators.tex

\subsection{Humans: They Often Disagree}\label{sec:annotators}

The final and possibly most problematic issue with system evaluation lies in the creation of the evaluation data itself. 
Common evaluation methodology assumes that there exists a single ground-truth for evaluation. 
This is a great oversimplification. 
We argue that evaluation should be conducted 
with reference to 
different groups of annotators 
to go beyond a one-dimensional performance score, to reflect multiple possible `truths'.


A great deal depends on how annotators are instructed to produce the data. 
It is well-known that human annotation quality may suffer from errors resulting from lack of attention given to the task, both by annotators themselves and by the annotation managers, often resulting from 
the need to drive annotation costs down \citep{mishra2021decides}.
Importantly, however, human label variation does not always reflect poor annotation. 
Label variation can also result from stimulus characteristics or the context in which annotation occurs, 
including factors like the identity of the annotators, their background, and world knowledge. 
\citet{plank2022problem} identifies three main reasons for human label variation, namely annotator disagreement, subjectivity (multiple possible perspectives) and underspecfication (multiple plausible answers). 
While subjectivity (e.g., due to cultural differences) is a clear issue in tasks like 
hate speech detection \cite{davani2021dealing}, inherent disagreements, ambiguous sentence meaning, underspecification in guidelines and annotator behavior have been identified not only in fine-grained Word Sense Disambiguation tasks \cite{navigli-2009}, but even in NLI \cite{pavlick-kwiatkowski-2019-inherent,zhang-de-marneffe-2021-identifying,jiang2022investigating}. 

While the standard approach for training and evaluating NLP systems is to use a single gold label for each example, a growing body of work deals with multiple labels by varying model training in various ways: 
different aggregation methods \citep{paun-etal-2018-comparing}, training on the distributional labels \citep{potts2020dynasent}, learning from agreement signals \citep{plank-etal-2014-learning}, or modeling the annotators \citep{geva-etal-2019-modeling,sap-etal-2022-annotators,gordon2022jury}.
Recently, \citet{basile-etal-2021-need} proposed extending this approach to evaluation. 
Fully benefiting from this extension requires releasing annotator characteristics labels \citep{prabhakaran-etal-2021-releasing}, including socio-demographic information, 
and carefully documenting the annotation process \cite{gebru2018datasheets,bender-friedman-2018-data,geiger2020garbage}.

Annotator disagreement often results from differences across individuals -- not just in NLP but also in fields such as cognitive science \citep{levinson2012original} and psycholinguistics \citep{kidd2018individual}. This phenomenon is often underestimated, since experiments 
tend to focus on a homogeneous sub-sample of the human population \citep{henrich2010weirdest}.
Annotators have different natural biases 
\citep{reidsma-op-den-akker-2008-exploiting}, and models often learn annotator-specific signals that are not generalizable \citep{geva-etal-2019-modeling}, including 
opinion, personality \citep{sap-etal-2022-annotators} and culture \cite{hershcovich-etal-2022-challenges}, but also different interpretation of guidelines  
\citep{hansen-sogaard-2021-guideline,parmar2022don}.
To deal with subjectivity, \citet{rottger-etal-2022-two} recently introduced two contrasting data annotation paradigms: the descriptive and prescriptive ones. While the former encourages annotator subjectivity by capturing and modelling different beliefs, the latter, instead, discourages it and enforces annotators to encode one specific belief, formulated in the annotation guidelines. Depending on the downstream application of the dataset, one paradigm can be more suitable than the other, but neither paradigm is inherently superior. However, dataset annotators should explicitly aim for one of the two paradigms to facilitate the intended downstream use of their dataset, and to document, for the benefit of others, how exactly their dataset was annotated.

In conclusion, without 
more attention to the ``science of annotation'', the methodological laxity in today's 
dataset creation 
will continue to foster 
inaccurate estimations of human performance. 

%% file: explanations.tex
\section{Humans Can Explain Their Answers}
\label{sec:explanations}

When performing language tasks, humans are capable of explaining why they provided a given answer. Thus, when models are claimed to attain 
human-level language understanding, we can reasonably expect to be able to elicit explanations from them. 
This has proven highly challenging, however, which casts further doubts on such claims. 

\paragraph{Why do we need explanations?}
At the level of an individual problem instance, explanations can help users assess whether to trust a given answer. At the level of a system, they help regulators and the general public to assess whether, or in what contexts, a system is safe to use, e.g.\ by uncovering unwanted biases or by revealing that the system relies on outdated knowledge. In the context of this paper, explanations can help NLP researchers understand the behaviour of their systems, e.g.\ to make sure that models are right for the right reasons \cite{mccoy-etal-2019-right,niven-kao-2019-probing}, or to uncover some of the shortcuts that the model may have learned \cite{geirhos2020shortcut}, as discussed in \S\ref{sec:right-for-wrong}. Indeed, the absence of explanations can lead researchers astray. For example, in a prize-winning paper, \citet{kaushik-lipton-2018-much} analysed several state-of-the-art QA systems 
and found that they simply classified the best matching answer using their pre-stored knowledge about each question candidate, without performing any `reading'. None of the papers in which these QA systems were introduced had considered this possibility. 

\paragraph{What are the challenges?}
While the importance of explanations is well-understood, progress has been hampered by various issues. One issue is that the evaluation of system-generated explanations is hard to automate (\S\ref{sec:automatic-evaluation}). Another issue is that it is not always clear what form the explanations should take. For tasks such as sentiment classification, it may be sufficient to highlight which words from the input text have mostly affected a given prediction. However, for NLI and QA, providing informative explanations can be challenging, even for humans. This can be observed by inspecting datasets that include human explanations \cite{camburu2018snli,rajani-etal-2019-explain,aggarwal-etal-2021-explanations}. 
Finally, system-generated explanations are typically not faithful, i.e.\ they do not necessarily reflect the process used by the model. For instance,
\citet{camburu-etal-2020-make} found that models can generate contradictory explanations for a given input. 

%% file: recommendations.tex
\section{Recommendations}

Based on the findings of the previous sections, we argue that current claims regarding superhuman performance are not adequately grounded, leading to unjustified hype. Here we provide a set of recommendations aimed at making comparisons between humans and machines fairer and more reliable.


\paragraph{Do not favor machines against humans}

Various actions can be taken to set a level playing field between humans and machines, so as  
to provide a more realistic sense of their actual performance:

\begin{enumerate} 
    \item \textbf{Avoid using the same documents for training and evaluation}  (\S\ref{sec:right-for-wrong}): 
    in fact, using the same documents 
    inherently reduces discrepancies across splits \citep{gorman-bedrick-2019-need}, encouraging models to learn specific idiosyncrasies that appear in both \cite{mccoy-etal-2019-right}. 
    \item \textbf{Balance easy and hard test set items} (\S\ref{sec:monolothic}), so as to report accuracies and enable analyses based on their difficulty level.
    \item \textbf{Occasionally refresh test sets}  (\S\ref{sec:overview}), as suggested by recent trends in adversarial evaluation \cite{kiela-etal-2021-dynabench}. 
    \item \textbf{Complement automatic evaluations with human judgements} (\S\ref{sec:automatic-evaluation}), so as to compare systems with humans on facets of language use that cannot be evaluated automatically.
    \item \textbf{Adequately train and motivate humans} (\S\ref{sec:issues}), aiming to increase the quality of human annotations through a solid training process and higher pay, in a sense mimicking the effort taken in improving systems.
\end{enumerate}

\paragraph{Make human performance evaluation transparent and reproducible}

We suggest carrying out an effort similar to systems' reproducibility for evaluating humans as well, including:

\begin{enumerate} 
    \item \textbf{Document the annotator pool composition} (\S\ref{sec:issues}), by explicitly answering the following questions: how many annotators were hired? Following what process? What is their cultural background, nationality, languages and areas of expertise?  What is their hourly pay rate? 
    \item \textbf{Specify the annotation process} (\S\ref{sec:issues} and \S\ref{sec:annotators}): it is important to state how many annotators were assigned to each instance, the training process they underwent, the guidelines they received (and how such guidelines were fine-tuned), and the metrics used to compute the overall human performance (averaging individual scores, majority voting, etc.). 
    \item \textbf{Provide individual annotations} (\S\ref{sec:annotators}): this allows recalculation of overall human performance whenever new metrics are tried, identifying the best metrics, calculating the scores of the best and worst annotators, the gap between the two, and the correlation between metrics and individual annotators -- all aspects that the annotation community has long advocated. Importantly, the availability of individual answers, combined with the annotators' profiles, opens the door to deeper investigations about why and when humans disagree. 


\end{enumerate}

\paragraph{Increase annotation accountability} Multiple measures can be implemented to make both systems and benchmarks more reliable, transparent and informative:

\begin{enumerate} 
    \item \textbf{Include explanations in your benchmark} (\S\ref{sec:explanations}):
    requiring annotators to provide the rationale behind their choices implicitly enforces them to devote more attention to the annotation task, thus yielding higher-quality and more consistent annotations. 
    Moreover, annotators' explanations can be used to study subjectivity, and discover (and mark) ambiguous instances.
    \item \textbf{Let systems produce explanations} (\S\ref{sec:explanations}): before claiming superhuman  performance, it is important that, similarly to humans, systems can explain the inferences behind their predictions. This is key both for increasing systems' credibility and for discovering their limitations. However, it is not impossible that a system will produce the right answer with the wrong explanation, or vice versa. For this reason, we believe that a system must be able to provide explanations that support its answers without knowing that answer a priori,  inferring 
    the answer based on its knowledge.  
\end{enumerate}

%% file: conclusions.tex
\section{Conclusions}

We have discussed the distressing tendency of many NLP researchers to claim superhuman performance for their systems, and outlined why such claims are not (yet) grounded. We identified problems with evaluation data, evaluation measures and methodology, system understandability, and the human creation of data, all of which contribute to our conclusion.  

As a final remark, with this paper we hope to make the reader more suspicious and rigorous when claims about ``superhuman'' performance are made, and, more importantly, 
to incentivize benchmark creators to address current limitations and design more solid and transparent benchmarks that will advance our scientific understanding of NLP systems and humans. 

%% file: limitations.tex
\section{Limitations}
In this paper, we have unearthed a variety of problems present in current evaluation benchmarks that favor systems over humans, or that simply make such comparisons unfair. 
We conclude that there is no real evidence to claim that today's language models possess superhuman performance. 
However, without empirical results obtained under the right setups, we cannot even claim the opposite, namely that humans are still better than systems.
We leave such demonstrations for future work.


Additionally, while a good portion of the NLP research effort is devoted to natural language generation (NLG) tasks (which includes MT), here we provide only some pointers to NLG/MT. Indeed, as discussed in Section \ref{sec:automatic-evaluation}, these problems exist in the NLG universe as well, but, due to space constraints, we limit our analysis to NLU tasks.

%% file: acknowledgements.tex
\section*{Acknowledgments}
    This work grew out of a brainstorming session held at the Rome Workshop on Ten Years of BabelNet in July 2022.\footnote{\url{http://mousse-project.org/events/event-a5f3r5.html}}
    \vspace{0.1cm}
\begin{center}
\noindent
    \begin{minipage}{0.1\linewidth}
        \begin{center}
            \includegraphics[scale=0.2]{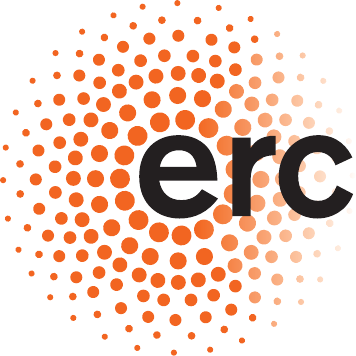}
        \end{center}
    \end{minipage}
    \hspace{0.01\linewidth}
    \begin{minipage}{0.70\linewidth}
         We gratefully acknowledge the support of the ERC Consolidator Grant MOUSSE No.\ 726487 under the European Union's Horizon 2020 research and innovation programme, and the support of the PNRR MUR project PE0000013-FAIR.
    \end{minipage}
    \hspace{0.01\linewidth}
    \begin{minipage}{0.1\linewidth}
        \begin{center}
            \includegraphics[scale=0.08]{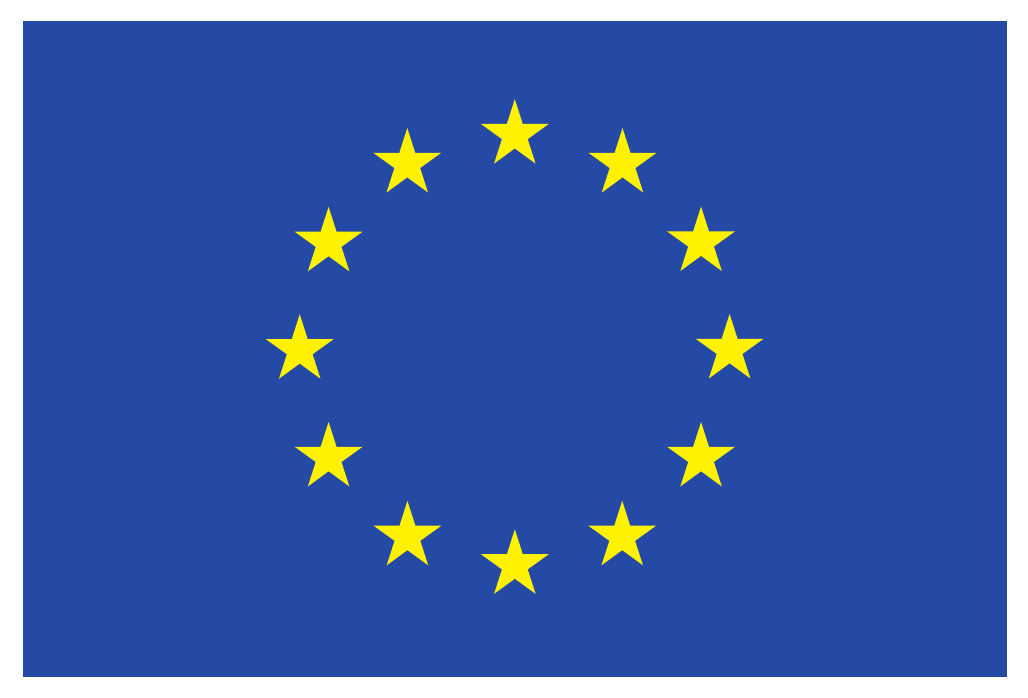}
        \end{center}
    \end{minipage}\\
\end{center}
\vspace{0.2cm}
This work has been carried out while Simone Tedeschi was enrolled in the Italian National Doctorate on Artificial Intelligence run by Sapienza University of Rome. The contribution of Jan Haji\v{c} has been supported by the LUSyD project No. GX20-16819X, funded by the Czech Science Foundation, and has used resources provided by the LRI LINDAT/CLARIAH-CZ, project LM2023062 funded by the M\v{S}MT CR.
The DFKI contribution to this work is supported by the LT-BRIDGE project, which has received funding from the European Union’s Horizon 2020 Research and Innovation Programme under Grant Agreement No. 952194. Rico Sennrich was funded by the Swiss National Science Foundation (grant no.~176727).

%% file: appendix-errors.tex

\section{Ground Truth Data Quality}\label{appendix:ground-truth-quality}
In Table \ref{tab:errors}, we reported one problematic example for each of the most crucial tasks in the SuperGLUE benchmark. Here we comment on those examples and provide additional problematic cases which we identified by manually inspecting the datasets. Some of these cases appear recurrently.

\begin{table*}[t]
\resizebox{\linewidth}{!}{%
\begin{tabular}{ll}
\toprule
\NameEntry{\textbf{Passage:} \textit{Columbidae -- The distinction between ``doves'' and ``pigeons'' is not consistent. In modern everyday speech, as opposed to scientific usage or formal usage, ``dove'' frequently indicates a pigeon that is white or nearly white. However, some people use the terms ``dove'' and ``pigeon'' interchangeably. In contrast, in scientific and ornithological practice, ``dove'' tends to be used for smaller species and ``pigeon'' for larger ones, but this is in no way consistently applied. Historically, the common names for these birds involve a great deal of variation between the terms. The species most commonly referred to as ``pigeon'' is the species known by scientists as the rock dove, one subspecies of which, the domestic pigeon, is common in many cities as the feral pigeon.}}\\
\\
\\
\\
\\
\\
\textbf{Question:} \textit{is there a difference between doves and pigeons} \quad \textbf{Answer:} \textsc{False}\\
\toprule
\NameEntry{\textbf{Passage:} \textit{Redback spider -- The redback is one of the few spider species that can be seriously harmful to humans, and its liking for habitats in built structures has led it to being responsible for a large number of serious spider bites in Australia. Predominantly neurotoxic to vertebrates, the venom gives rise to the syndrome of latrodectism in humans; this starts with pain around the bite site, which typically becomes severe and progresses up the bitten limb and persists for over 24 hours. Sweating in localised patches of skin occasionally occurs and is highly indicative of latrodectism. Generalised symptoms of nausea, vomiting, headache, and agitation may also occur and indicate severe envenomation. An antivenom has been available since 1956. There have been no deaths directly due to redback bites since its introduction, however Isbister et al. have suggested patients for whom antivenom is considered should be fully informed ``there is considerable weight of evidence to suggest it is no better than placebo'', and in light of a risk of anaphylaxis and serum sickness, ``routine use of the antivenom is therefore not recommended''. As of the 2013 (updated 2014) edition of the Snakebite \& Spiderbite Clinical Management Guidelines from NSW HEALTH (latest available in 2017), Red-back spider bites were considered not life-threatening but capable of causing severe pain and systemic symptoms that could continue for hours to days.
}}\\
\\
\\
\\
\\
\\
\\
\\
\\
\\
\\
\\
\textbf{Question:} \textit{can a red back spider bite kill you} \quad \textbf{Answer:} \textsc{True}\\
\toprule
\NameEntry{\textbf{Passage:} \textit{Mobile phones in prison -- In most prisons, inmates are forbidden from possessing mobile phones due to their ability to communicate with the outside world and other security issues. Mobile phones are one of the most smuggled items into prisons. They provide inmates the ability to make and receive unauthorized phone calls, send email and text messages, use social media, and follow news pertaining to their case, among other forbidden uses.}}\\
\\
\\
\\
\textbf{Question:} \textit{are you allowed to have your phone in prison} \quad \textbf{Answer:} \textsc{False}\\
\toprule
\NameEntry{\textbf{Passage:} \textit{Vena amoris -- Vena amoris is a Latin name meaning, literally, ``vein of love''. Traditional belief established that this vein ran directly from the fourth finger of the left hand to the heart. This theory has been cited in western cultures as one of the reasons the engagement ring and/or wedding ring was placed on the fourth finger, or ``ring finger''. This traditional belief is factually inaccurate as all the fingers in the hand have a similar vein structure.}}\\
\\
\\
\\
\textbf{Question:} \textit{is it true that the ring finger is connected to the heart} \quad \textbf{Answer:} \textsc{False}\\
\toprule
\NameEntry{\textbf{Passage:} \textit{Substitute (association football) -- Most competitions only allow each team to make a maximum of three substitutions during a game and a fourth substitute during extra time, although more substitutions are often permitted in non-competitive fixtures such as friendlies. A fourth substitution in extra time was first implemented in recent tournaments, including the 2016 Summer Olympic Games, the 2017 FIFA Confederations Cup and the 2017 CONCACAF Gold Cup final. A fourth substitute in extra time has been approved for use in the elimination rounds at the 2018 FIFA World Cup, the UEFA Champions League and the UEFA Europa League. Each team nominates a number of players (typically between five and seven, depending on the competition) who may be used as substitutes; these players typically sit in the technical area with the coaches, and are said to be ``on the bench''. When the substitute enters the field of play it is said they have come on or have been brought on, while the player they are substituting is coming off or being brought off.}}\\
\\
\\
\\
\\
\\
\\
\\
\\
\textbf{Question:} \textit{can a player be substituted twice in football} \quad \textbf{Answer:} \textsc{True}\\
\bottomrule
\end{tabular}%
}
\caption{Additional problematic instances we have found in the BoolQ dataset.
}\label{tab:boolq-errors}
\end{table*}

\begin{table*}[!h]
\resizebox{\linewidth}{!}{%
\begin{tabular}{ll}
\toprule 
\NameEntry{\textbf{Paragraph:} \textit{Amateur tennis star Guy Haines wants to divorce his vulgar and unfaithful wife Miriam , so he can marry the elegant and beautiful Anne Morton , daughter of a senator .  While on a train to meet Miriam , Haines meets Bruno Anthony , a forward stranger who recognizes Guy from gossip items in the newspapers that detail his marital problems .  During lunch in Bruno's compartment , Bruno tells Guy about his idea for the perfect `` Criss-cross '' murder : he will kill Miriam and in exchange , Guy will kill Bruno's father .  Since both are strangers , otherwise unconnected , there is no identifiable motive for the crimes , Bruno contends , hence no suspicion .  Guy hurriedly leaves the compartment but leaves Bruno thinking he has agreed to the deal .  Guy accidentally leaves his cigarette lighter behind , a gift from Anne to Guy , Which Bruno pockets .  Bruno heads to Guy's hometown of Metcalf and follows Miriam and her two beaux to an amusement park , where he briefly illuminates her face with Guy's lighter , then strangles her to death .  Guy's problems begin when his alibi an inebriated college professor on the same train as Guy can not remember their meeting .  But they increase exponentially when Bruno makes repeated appearances into Guy's life as he seeks to remind Guy that he is now obliged to kill Bruno's father , according to the bargain he thinks they struck on the train .  Bruno sends Guy the keys to his house , a map to his father's room , and a pistol .  Soon after , Bruno appears at a party at Senator Morton's house and hobnobs with the guests , much to Guy's apprehension and Anne's increasing suspicion.}}\\
\\
\\
\\
\\
\\
\\
\\
\\
\\
\\
\\
\\
\textbf{Question:} \textit{Who are the two that Guty and Bruno are planning to murder?} \quad \textbf{Candidate Answers}: \textit{Bruno's father} (\textsc{True}), \\ \textit{Guy's father} (\textsc{False}), \textit{Bruno's wife} (\textsc{False}), \textit{Miriam and Bruno's father} (\textsc{True}), \textit{Guy's wife} (\textsc{True}), 
$\dots$\\
\toprule 
\NameEntry{\textbf{Paragraph:} \textit{Albert Bandura OC (/baen'dU@r@/; born December 4, 1925) is a psychologist who is the David Starr Jordan Professor Emeritus of Social Science in Psychology at Stanford University. For almost six decades, he has been responsible for contributions to the field of education and to many fields of psychology, including social cognitive theory, therapy and personality psychology, and was also influential in the transition between behaviorism and cognitive psychology. He is known as the originator of social learning theory and the theoretical construct of self-efficacy, and is also responsible for the influential 1961 Bobo doll experiment. Social learning theory is how people learn through observing others. An example of social learning theory would be the students imitating the teacher. Self-efficacy is "The belief in one's capabilities to organize and execute the courses of action required to manage prospective situations." To paraphrase, self-efficiacy is believing in yourself to take action. The Bobo Doll Experiment was how Albert Bandura studied aggression and non-aggression in children. A 2002 survey ranked Bandura as the fourth most-frequently cited psychologist of all time, behind B. F. Skinner, Sigmund Freud, and Jean Piaget, and as the most cited living one. Bandura is widely described as the greatest living psychologist, and as one of the most influential psychologists of all time. In 1974 Bandura was elected to be the Eighty-Second President of the American Psychological Association (APA). He was one of the youngest president-elects in the history of the APA at the age of 48. Bandura served as a member of the APA Board of Scientific Affairs from 1968 to 1970 and is well known as a member of the editorial board of nine psychology journals including the Journal of Personality and Social Psychology from 1963 to 1972. At the age of 82, Bandura was awarded the Grawemeyer Award for psychology.}}\\
\\
\\
\\
\\
\\
\\
\\
\\
\\
\\
\\
\\
\\
\\
\\
\textbf{Question:} \textit{In what year was Bandura awarded the Grawemeyer Award for psychology.} \quad \textbf{Candidate Answers}: \textit{2010} (\textsc{False}), \\ \textit{2007} (\textsc{True}), \textit{2000} (\textsc{False}), \textit{2002} (\textsc{True})\\
\toprule 
\NameEntry{\textbf{Paragraph:} \textit{(CNN) -- German art collector Cornelius Gurlitt, whose nearly priceless collection was confiscated because it was suspected to contain pieces looted by the Nazis, died Tuesday and left the masterpieces to a Swiss museum. One day after Gurlitt's death at the age of 81, the Museum of Fine Arts Bern announced that Gurlitt had named it \"his unrestricted and unfettered sole heir.\" The news came as a surprise, the museum said Wednesday, because Gurlitt had never had any connection to it. The museum's directors are delighted at the news, they said in a statement, but also recognize that there are outstanding legal and ethical questions surrounding the collection. Gurlitt had undergone major heart surgery and was hospitalized for many weeks, his representative said in a statement. Gurlitt grabbed the attention of the art world when German prosecutors seized more than 1,200 paintings from his Munich apartment in 2012, including works by Picasso and Matisse. The collection was confiscated as part of an investigation into tax fraud, but then it was thought that some of the paintings may have been works that were looted by the Nazis. Just last month, part of the collection was returned to Gurlitt as part of a deal with Germany's cultural authorities and the Bavarian Justice Ministry. Under the agreement, works owned by Gurlitt that were not under suspicion were returned to him. Those suspected of being stolen were to be held securely while a task force investigates their provenance -- and will be returned to their original Jewish owners or their descendants if a claim is proven. Gurlitt's representative said that with the art collector's death, the investigation into the collection ceases. The court that was handling the investigation proceedings will now function as an estate court in the case.}}\\
\\
\\
\\
\\
\\
\\
\\
\\
\\
\\
\\
\\
\\
\textbf{Question:} \textit{How old was the art collector Cornelius Gurlitt when he died?} \quad \textbf{Candidate Answers}: \textit{At the age of 81} (\textsc{True}),\\ \textit{80} (\textsc{False}), \textit{80 years old} (\textsc{True}), \textit{81} (\textsc{False})\\
\bottomrule
\end{tabular}%
}
\caption{Additional problematic instances in the Multi-Sentence Reading Comprehension (MultiRC) dataset.}\label{tab:multirc-errors}
\end{table*}

\begin{table}[t]
\resizebox{\columnwidth}{!}{%
\begin{tabular}{ll}
\toprule
\NameEntryTwo{\textbf{Premise:} \textit{{A: Your turn. B: Okay. Uh, I don't think they should abolish it.
}}}\\
\\
\textbf{Hypothesis:} \textit{they should abolish it}\\
\textbf{Entailment:} \textsc{False}\\
\toprule
\NameEntryTwo{\textbf{Premise:} \textit{The lunch trade had mostly disappeared so he wasn't hard to spot. He was at a window table but he was ignoring the river, being deep in conversation with a middle-aged man wearing a suit and a short sheepskin car coat with matching brown suede shoes. Even from this distance you could guess the guy's tailor was based in Dublin.
}}\\
\\
\\
\\
\\
\\
\textbf{Hypothesis:} \textit{the guy's tailor was based in Dublin} \\
\textbf{Entailment:} \textsc{True}\\
\toprule 
\NameEntryTwo{\textbf{Premise:} \textit{B: and, you know, they just love kittens. A: Yeah. B: They just are fascinated. A: Oh, yeah. B: So she doesn't know that this is a cat yet.}}\\
\\
\\
\textbf{Hypothesis:} \textit{this is a cat} \\
\textbf{Entailment:} \textsc{True}\\
\toprule
\NameEntryTwo{\textbf{Premise:} \textit{A: Well, actually, uh, A: I don't think I'm in the, uh, majority in Texas}}\\
\\
\textbf{Hypothesis:} \textit{she is in the majority in Texas} \\
\textbf{Entailment:} \textsc{False}\\
\toprule
\NameEntryTwo{\textbf{Premise:} \textit{B: Because too often, there can be extremism that hurts from any direction, regardless of whatever you're arguing or concerned about. A: Yeah.  Right. Yeah, I know, you're right, they would lobby that and I see that, and that's why, you know, I'm like, okay, what's my role in this thing,, you know, what's my part, B: Yeah. A: because I don't think the system is going to get fixed.}}\\
\\
\\
\\
\\
\\
\textbf{Hypothesis:} \textit{the system is going to get fixed.}\\
\textbf{Entailment:} \textsc{False}\\
\bottomrule
\end{tabular}%
}
\caption{Additional problematic instances we have found in the CommitmentBank (CB) dataset.
}\label{tab:cb-errors}
\end{table}

\begin{table}[h]
\resizebox{\columnwidth}{!}{%
\begin{tabular}{ll}
\toprule 
\NameEntryTwo{\textbf{Premise:} \textit{Compuware claims that Allan Tortorice and Jim Hildner were among several former employees who revealed trade secrets after they moved to IBM.}}\\
\\
\\
\textbf{Hypothesis:} \textit{Trade secrets were stolen.} \\
\textbf{Entailment:} \textsc{False}\\
\toprule
\NameEntryTwo{\textbf{Premise:} \textit{{It has been observed that in those countries of the world where capital punishment is still in operation, the crime rate, especially murder, is distinctively low in comparison to countries where capital punishment has been discarded.
}}}\\
\\
\\
\\
\textbf{Hypothesis:} \textit{Capital punishment is a deterrent to crime.} \\
\textbf{Entailment:} \textsc{True}\\
\toprule
\NameEntryTwo{\textbf{Premise:} \textit{A farmer who was in contact with cows suffering from BSE -- the so-called mad cow disease -- has died from what is regarded as the human form of the disease.}}\\
\\
\\
\NameEntryTwo{\textbf{Hypothesis:} \textit{Bovine spongiform encephalopathy is another name for the "mad cow disease"}} \\
\\
\textbf{Entailment:} \textsc{True}\\
\toprule
\textbf{Premise:} \textit{The girl was found in Drummondville.}\\
\textbf{Hypothesis:} \textit{Drummondville contains the girl.} \\
\textbf{Entailment:} \textsc{False}\\
\toprule
\NameEntryTwo{\textbf{Premise:} \textit{The official visit of the Argentine minister marks a further step in the normalisation of UK-Argentine relations.}}\\
\\
\NameEntryTwo{\textbf{Hypothesis:} \textit{Relations between Argentina and Great Britain are growing more cooperative.}}\\
\\
\textbf{Entailment:} \textsc{False}\\
\bottomrule
\end{tabular}%
}
\caption{Some problematic instances we have found in the Recognizing Textual Entailment (RTE) dataset.
}\label{tab:rte-errors}
\end{table}

\begin{table}[!t]
\resizebox{\columnwidth}{!}{%
\begin{tabular}{ll}
\toprule
\textbf{Context 1:} \textit{I tried to make a call, but the \underline{line} was dead.}\\ \textbf{Context 2:} \textit{A dedicated \underline{line}.}\\
\textbf{Sense Match:} \textsc{True}\\
\toprule
\textbf{Context 1:} \textit{The author gives a depressing \underline{picture} of life in Poland.}\\ \textbf{Context 2:} \textit{He had no clear \underline{picture} of himself or his world.}\\
\textbf{Sense Match:} \textsc{False}\\
\toprule
\textbf{Context 1:} \textit{Instant replay caused too long a \underline{delay}.}\\ \textbf{Context 2:} \textit{The \underline{delay} before the echo of a sound.}\\
\textbf{Sense Match:} \textsc{False}\\
\toprule
\textbf{Context 1:} \textit{\underline{Stop} a car.}\\ \textbf{Context 2:} \textit{\underline{Stop} the thief.}\\
\textbf{Sense Match:} \textsc{True}\\
\toprule
\textbf{Context 1:} \textit{\underline{Fall} asleep.}\\ \textbf{Context 2:} \textit{She \underline{fell} to pieces after she lost her work.}\\
\textbf{Sense Match:} \textsc{True}\\
\bottomrule
\end{tabular}%
}
\caption{Additional problematic instances we have found in the Word-in-Context (WiC) dataset.
}\label{tab:wic-errors}
\end{table}

\paragraph{BoolQ} The example in Table \ref{tab:errors} is blatantly wrong, as it explicitly says that shower gel is an \textit{effective and perfectly acceptable substitute to shampoo}, hence the label should be \textsc{False}. We provide more errors in Table \ref{tab:boolq-errors}. Specifically, we believe that some of these examples are wrongly annotated, ambiguous, or highly misleading. 
In the first example, from the premise, it seems that \textit{some scientists and ornithologists differentiate between doves and pigeons}, so the answer might be subjective, and therefore ambiguous. 
In the second example, instead, it seems there is no evidence that a red back spider bite can kill a human being, but the answer is \textsc{True}. 
Similarly to the first case, in the third example the premise states that \textit{in most prisons} possession of mobile phones is not allowed, thus the answer might change depending on the prison. 
In the fourth example, the fact that \textit{all the fingers have a similar vein structure} does not mean that \textit{the ring finger is not connected to the heart}, on the contrary, this reinforces the hypothesis. 
Finally, while two or more players can be substituted in a football game, the same player cannot be substituted twice. 

\paragraph{CommitmentBank} In the CB example reported in Table \ref{tab:errors} we have that A does not know whether \textit{they've ever made a movie} and, indeed, asks if B thinks they have. Therefore, we cannot conclude that \textit{the movie was never made}, and the answer should be \textsc{Neutral}. 
By inspecting the dataset, we discovered that its instances follow standard patterns that can be easily learned by machines, but, at the same time, confuse humans. Indeed, most of the time, the entailment is \textsc{True} when a fragment of the hypothesis appears (as an exact match) in the premise (see the second and third examples in Table \ref{tab:cb-errors}). To the contrary, the entailment is \textsc{False} when the same text fragment appears negated either in the premise or in the hypothesis, e.g. preceded by \textit{don't think} or similar, standard constructs (see the first, fourth and fifth examples in Table \ref{tab:cb-errors}). However, as argued before, the mere fact of not thinking that a thing is true does not necessarily imply that thing is not true.

\paragraph{MultiRC} Regarding the MultiRC example of Table \ref{tab:errors}, in this case, the error is in the candidate answers. Specifically, two candidate answers are equivalent, i.e. \textit{Mass of the object} and \textit{The object's mass}, but they are labeled differently, namely with \textsc{True} and \textsc{False} tags, respectively. In Table \ref{tab:multirc-errors} we provide additional errors for this task.
Specifically, in the first example, the question explicitly asks \textit{``Who are the \underline{two} that Guty and Bruno are planning to murder?''}, but the possible answers are i) \textit{Miriam and Bruno's father}, ii) \textit{Bruno's father} and iii) \textit{Guy's wife}. 
Although, by design, MultiRC creators explicitly state that multiple answers can be correct, answers are judged independently, so it would not be valid to form a correct answer by combining ii) and iii). 
These cases are very frequent in MultiRC and might have negatively affected human performance. Furthermore, typos in the questions and/or paragraphs (i.e. \textit{Guty}, in this case) might have further limited their scores.

In the second example, the ground truth answers are ``2002'' and ``2007''. However, while ``2007'' can be inferred by adding 82 years (i.e. the age at which Albert Bandura received the Grawemeyer award) to his birth date (i.e. 1925), ``2002'' is a wrong answer. Indeed, the paragraph says that ``\textit{A 2002 survey ranked Bandura as the fourth
most-frequently cited psychologist of all time}'', but there is no evidence that he received the award in 2002.

Finally, in the third example, from the paragraph, it is clear that the German art collector Cornelius Gurlitt passed away at the age of 81. However, there are three errors in the possible answers for this entry. First, ``\textit{At the age of 81}'' and ``\textit{81}'' are labeled as \textsc{True} and \textsc{False}, respectively. Second, ``\textit{80 years old}'' is labeled as \textsc{True}, hence contradicting the first answer. Finally, ``\textit{80}'' is labeled as \textsc{False} further contradicting the penultimate answer.

\paragraph{RTE} In the RTE example (Table \ref{tab:errors}), the specific premise regarding \textit{Pacific countries} is not sufficient to entail the general hypothesis, thus the answer should be \textsc{False}. We provide more examples in Table \ref{tab:rte-errors}. In particular, in some of them, we believe that the label is incorrect (examples 1, 3 and 5), or at least highly misleading, while in some others we think that not enough information is provided to entail the hypothesis (examples 2 and 4).

\paragraph{WiC} In the WiC example provided in Table \ref{tab:errors}, the word \textit{criticism} is used with the same meaning in the two contexts, namely \textit{disapproval expressed by pointing out faults or shortcomings} according to WordNet. We provide additional ambiguous or wrongly annotated examples in Table \ref{tab:wic-errors}. 

By inspecting the WiC dataset, it is immediately apparent that, in many negative examples, the semantic gap between the meanings of the same lemma in the two contexts is very narrow. Although such cases are difficult even for machines, we posit that for humans (especially if sense distinctions are not provided and annotators are not lexicographers, as in WiC) they are way more difficult.

\paragraph{SQuAD}
For the SQuAD dataset, studies about errors in the annotations have already been performed by \citet{rodriguez-etal-2021-evaluation} through automatic error detection methods. Specifically, they annotated SQuAD items by discriminability, difficulty, and Item Response Theory (IRT) prediction errors, and discovered that items with negative discriminability, or where IRT’s prediction is wrong, have a much higher rate of annotation error, i.e. they are often ``flawed'' or ``wrong''.
We believe that tools for error detection \cite{10.1162/coli_a_00464} should play a key role in the improvement of existing benchmarks and in the creation of new ones.

Finally, still related to the topic of wrongly-annotated or ambiguous instances in the datasets, \citet{nangia-bowman-2019-human} performed an interesting study on the GLUE benchmark. In order to investigate the effect of these instances, they looked at human performance when there is 5-way annotator
agreement. Using unanimous agreement has the effect of filtering out examples for
which: i) the annotation guidelines supplied do not provide clear advice, and ii) humans understand the expectations of the task but find the example genuinely difficult or uncertain. They discovered that this widened the gap between humans and systems by more than 3 points on average, hence confirming the hypothesis that humans were often penalized by unclear guidelines or other factors. Even more interestingly, they found that in some tasks, when systems are evaluated on the unanimous subsets they obtain lower scores compared to those obtained on the entire test sets containing wrong or ambiguous instances, hence suggesting that systems had learned specific idiosyncrasies appearing in both training and test sets (Section \ref{sec:right-for-wrong}).

\section{Apples and Oranges}\label{app:apples-and-oranges}
In Section \ref{sec:issues}, the first issue that we pointed out was that, on almost all SuperGLUE tasks, humans and machines are evaluated on different test sets (i.e. on a small subset vs. the full test set). Here, we provide more details (see Table \ref{tab:percentages}). Specifically, it can be observed that only 3 out of 10 tasks are fully annotated by humans, while for the remaining 7 tasks only a small portion is annotated, ranging from 3\% to 40\% of the full dataset size. 

\begin{table}[t]
\centering
\resizebox{0.75\columnwidth}{!}{%
    \begin{tabular}{lrrr}
        \toprule
        \textbf{Task} & \textbf{H} & \textbf{S} & \textbf{$\%$} \\
        \midrule
        AX-b & 100 & 1104 & 9.05\%\\
        AX-g & 100 & 356 & 28.08\%\\
        BoolQ & 100 & 3245 & 3.08\%\\
        CB & 100 & 250 & 40.00\%\\
        COPA & 100 & 500 & 20.00\%\\
        MultiRC & 166 & 166 & 100.00\%\\
        RTE & 500 & 3000 & 16.67\%\\
        ReCoRD & 10000 & 10000 & 100.00\%\\
        WSC & 146 & 146 & 100.00\%\\
        WiC & 300 & 1400 & 21.42\%\\
        \bottomrule
    \end{tabular}%
    }
    \caption{Comparison of the size of the test sets annotated by humans (H) and systems (S) in the various tasks. \% represents the ratio between H and S.}
    \label{tab:percentages}
\end{table}

\begin{table*}[htbp]
\resizebox{\linewidth}{!}{%
\begin{tabular}{llllrr}
\toprule
\textbf{WiC} & \textbf{Target} & \textbf{Context-1} & \textbf{Context-2} & \textbf{F} & \textbf{T}\\
\midrule
T & line (N) & I tried to make a call, but the \underline{line} was dead. & A dedicated \underline{line}. & 14 & 11 \\
F & love (N) & A mother's \underline{love} is not easily shaken.	& The theater was her first \underline{love}. & 16 & 9\\
F & work (V) & This dough does not \underline{work} easily. & \underline{Work} the phones. & 16 & 9\\
T & fall (V) & \underline{Fall} asleep. & She \underline{fell} to pieces after she lost her work. & 17 & 7\\
F & picture (N) & The author gives a depressing \underline{picture} of life in Poland. & He had no clear \underline{picture} of himself or his world. & 7	& 18\\
F & take (V) & Do you \underline{take} sugar in your coffee? & A reading was \underline{taken} of the earth's tremors. & 21 & 4\\
\bottomrule
\end{tabular}%
}
\caption{Results of the Copenhagen experiment. Context-1 and Context-2 provide two sentences in which a Target word $w$ appears. The WiC column specifies the label in the WiC dataset, indicating whether $w$ has the same meaning in the two contexts. Finally, the F and T columns represent the number of people who voted for F and T, respectively.}\label{tab:copenhagen}
\end{table*}

\section{The Copenhagen Experiment}\label{app:copenhagen-experiment}
In this Section, we report on an experiment that was conducted in Copenhagen at the Danish Language Technology Conference in 2022\footnote{\url{https://cst.ku.dk/kalender/sprogteknologisk-konference-2022/}}. The main goal was to verify the claim that contextual sentence examples from open lexical resources, such as those used to create the WiC dataset, i.e. WordNet, VerbNet and Wiktionary, ``constitute a reliable base for the construction of the dataset, as they are curated in a way to be clearly distinguishable across different senses of the word'' \cite{pilehvar-camacho-collados-2019-wic}. 
Based on this assumption, ``the [WiC dataset] annotators were not provided with knowledge from any external lexical resource'', and were asked to label each instance solely based on whether they thought the two occurrences of the word referred to the same meaning or not. 

We repeated the above annotation task by asking 25 
conference participants to provide ``true'' (T) and ``false'' (F) answers for the six instances in Table \ref{tab:copenhagen}. The results show a high degree of disagreement, suggesting that the above claim is not always valid, especially when subtle sense distinctions are involved. We posit that the presence of a certain amount of intrinsically debatable items hampers fair comparisons between humans and systems.

Indeed, in WSD evaluation tasks, where the granularity of senses is a key concern \cite{ijcai2021p593}, we advocate that the starting point for the task design should consist either of the warning made by many lexicographers that ``there is very little agreement about what word senses are or how broad their scope should be, and no definitive way of knowing when one sense ends and another begins'' \cite{atkins2008oxford}, or the one from the famous lexicographer, James Murray, that ``the best any lexicographer could hope for would be that readers would feel, on scanning a multisense dictionary entry, that this is not an unreasonable way of exhibiting the facts''. With large corpora and the latest advances in language modeling, we now have the possibility to measure differences between contexts in which words are used, and we need not and should not rely on made-up sentences from the times when corpora were not available at all. This is corroborated, for instance, by the inter-tagger agreement and the systems' results of the multilingual version of the WiC task, where sentences come from real text and dictionary definitions are used as a help for annotators \cite{martelli-etal-2021-semeval}.